\documentclass[binding=0.6cm,LaM,oneside]{sapthesis}

\usepackage{microtype}
\usepackage[italian,english]{babel}
\usepackage[utf8]{inputenx}
\usepackage{lettrine}
\usepackage{listings}
\usepackage{multirow}
\usepackage{subcaption,siunitx,booktabs}
\usepackage{graphicx}
\title{Pushing the envelope in deep visual recognition for mobile platforms}
\author{Lorenzo Alvino}
\IDnumber{1722362}
\course{Artificial Intelligence and Robotics}
\courseorganizer{Facolt\`{a} di Ingegneria dell'Informazione, Informatica e Statistica}
\AcademicYear{2016/2017}
\advisor{Prof. Barbara Caputo}
\copyyear{2017}
\authoremail{lorenzo.alvino@outlook.it}
\examdate{24 October 2017}
\examiner{Prof. Giuseppe De Giacomo} \examiner{Prof. Becchetti Luca} \examiner{Prof. Bonomi Silvia}  \examiner{Prof. Grisetti Giorgio}  \examiner{Prof. Pirri Fiora} \examiner{Prof. Vitaletti Andrea}  \examiner{Prof. Wollan Paul Joseph}

\begin{document}

\frontmatter
\maketitle
\dedication{To my mother,\\who was my\\strength and courage\\
\bigskip\bigskip\bigskip\bigskip\bigskip\bigskip\bigskip\bigskip\bigskip\bigskip\bigskip\bigskip To Chiara,\\who truly\\believed in me}

\tableofcontents
\begin{abstract}
Image classification is the task of assigning to an input image a label from a fixed set of categories. It's one of the core problems in computer vision that, despite its simplicity, has a large variety of practical applications.\\
In a typical situation, a classification model takes as input a single image and assigns a single label or a probability distribution over the different labels as output.
One of the most important applicative fields is that of robotics, in particular the needing of a robot to be aware of what's around and the consequent exploitation of that information as a benefit for its tasks. In this work we consider the problem of a robot that enters a new environment and wants to understand visual data coming from its camera, so to extract knowledge from them. As main novelty we want to overcome the needing of a physical robot, as it could be expensive and unhandy, so to hopefully enhance, speed up and ease the research in this field. That's why we propose to develop an application for a mobile platform that wraps several deep visual recognition tasks.\\
First we deal with a simple Image classification, testing a model obtained from an AlexNet trained on the ILSVRC 2012 dataset. Several photo settings are considered to better understand which factors affect most the quality of classification.
For the same purpose we are interested to integrate the classification task with an extra module dealing with segmentation of the object inside the image. In particular we propose a technique for extracting the object shape and moving out all the background, so to focus the classification only on the region occupied by the object.
Another significant task that is included is that of object discovery. Its purpose is to simulate the situation in which the robot needs a certain object to complete one of its activities. It starts searching for what it needs by looking around and trying to understand the location of the object by scanning the surrounding environment. Finally we provide a tool for dealing with the creation of customized task-specific databases, meant to better suit to one's needing in a particular vision task.
\end{abstract}

\mainmatter
\chapter{Overview}
\lettrine[lines=2, findent=3pt, nindent=0pt]{N}{}owadays it seems as though almost everyone owns a smart phone, even if the first time in which they came to play, starting to become popular between common people wasn't so long ago. Anyway their capabilities made huge steps in a handful of years and they shortly evolved in the do-all devices, that society takes for granted, of today. With their advanced computing capabilities and other features, smart phones became popular very quickly, so that we can state that almost every human being has got one and takes it as part of its life. In figure 1.1 a statistic about smart phone users is shown.

\bigskip
\begin{figure}[ht]
 \centering
 \captionsetup{justification=centering}
 \includegraphics{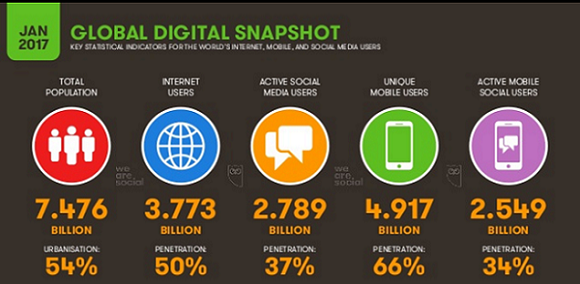}
 \caption{Statistics referred to mobile users in 2017\\ provided by We Are Social and Hootsuite}
\end{figure}
\bigskip

The spread of smart phones around the planet constitutes the key point in which mobile technology and computer vision applied to the field of robotics meet together. The main reason of our work is that having a mobile robot that gathers data relevant to some vision task and tests learned models for image classification could be tricky, uncomfortable and expensive. That's why we would like to create a mobile application that can take advantage of the diffusion of smart phones and the consequent great availability of cameras. The idea is to propose a mobile tool that can substitute robot's eyes to test learned models or to capture objects from a robot’s perspective. The system would allow to go around, taking pictures or videos of objects in the surrounding environment and use them for classification purposes to extract information that could be useful for other robotics tasks, such as navigation, obstacle avoidance, manipulation and so on.\\
To extract knowledge from visual data, a neural network is trained on a very big dataset containing many labeled images, referring to a thousand different object categories. The output of the training is a model that can be used to classify images. To measure its performance, in term of accuracy, it is evaluated on the image classification task \cite{classification}, taking into account several settings that could occur in a typical situation. The aspects concerning the action of taking a photo are multiple, such as different light conditions, object views, backgrounds and distances. They are all evaluated, in terms of model performance, to know which combination would be optimal for taking a photo to be classified.
To further enhance the quality of the model, we introduce a segmentation step before the classification. The idea is that, if the system is able to provide the model with an image that is the cleanest possible, then the label assignment will be more precise. For cleanness we mean extracting that portion of image containing the object of interest, removing all the background, or almost all of it, so to pull out the object, with its smooth boundary, in the most faithful way possible.\\
The object discovery task is considered as well. It consists in capturing a small scan of a scene to look for the presence of a particular object in it. The system takes as input frames extracted from the scan and an object label, to return the scene snippet containing the searched object. As an example, you could think about it as a robot that enters a certain environment and wants to check the presence of a certain object relevant to one of its tasks. Finally a shortcut to create customized databases is provided. The function allows one to create its own task-specific database of annotated images by scanning the objects of interest from different views.
\section{Outline}
The rest of the thesis is organized as follows:\\

In sections 1.2 - 1.5 we give an overview of the main computer vision problems the work relies on. General notions are given, together with typical examples related to this work, to have an insight into deep visual recognition.\\

In chapter 2 we mention the works related to this thesis, spanning the main fields treated in our project.\\

Chapter 3 points out the architecture that is used to train the model for image classification, as well as the data to perform the training. The segmentation module is described too, highlighting its main functions and components.\\

In chapter 4 we deeply describe the implementation choices and comment in detail the core of the code that is behind the scenes\\

Chapter 5 explains the setup up of the experiments, focusing attention on the main assumptions made before evaluation and the obtained results.\\

In chapter 6 we end the thesis with a conclusive recap and some mentions about the possible future work directions. 
\bigskip
\newpage
\section{Image Classification}
Image classification \cite{gallegoapplication} is one of the core problems in computer vision. It consists in assigning a category to an input image, choosing between a set of known labels. For a computer an image is represented by one large 3-dimensional array of numbers. Each number is an integer that ranges from 0 to 255, where the two extremities refer, respectively, to black and white color. The classification consists in turning this large amount of numbers into a single label, that is the output of the task.

\bigskip
\begin{figure}[ht]
 \centering
 \includegraphics[scale=0.8]{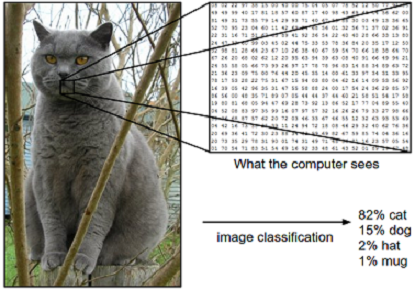}
 \caption{Example of image classification}
\end{figure}
\bigskip

A toy example of what happens is shown in the figure above. A picture of a cat is passed to a classification model which has to assign probabilities to 4 labels: cat, dog, hat, mug. In this case the cat image accounts for 248 pixels of width, 400 pixels of height and three channels, referring to the Red, Green and Blue colors (RGB). Counting the numbers that come to play as $width \times height \times channels$, it turns out that, even though we are dealing with a simple case, they are  297,600. The goal is to turn this quarter of a million numbers into a single label, in particular “cat”.\\
It might seem a trivial task for a human being, and in fact it is, but from a computer viewpoint things are more challenging. For a computer vision algorithm there are many  factors coming in, such as:
\begin{itemize}
\item \textbf{Viewpoint and scale.} The viewpoint of an object with respect to a certain camera pose could vary as the object itself could be oriented in many ways. For the scale two variation aspects must be considered: the scale that the object has in the image, that is the portion of space it occupies, as well as the scale of the object in the real world, meant as the difference in size of objects belonging to the same category.
\item \textbf{Illumination conditions}. The effect of different illumination conditions (natural, artificial, flash), could be really heavy on the pixel level, in particular in altering object colors.

\bigskip
\begin{figure}[ht]
 \centering
 \includegraphics[scale=0.8]{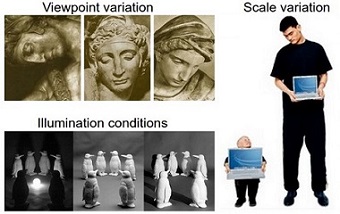}
 \caption{Visual challenges: viewpoint, scale, illumination}
\end{figure}
\bigskip

\item \textbf{Deformation.} An object could be deformable because it isn't made of a rigid material and so its shape can be altered. But deformation can also refer to moving parts of a model, for example body parts in a pedestrian detector, where the object of interest assumes different shapes as it moves over time.
\item \textbf{Occlusion and background clutter.} Depending on the type of surrounding environment, the object of interest could be occluded by other ones, even in a substantial way, rending the classification task much harder as few pixels are visible. Background cluttering must be carefully considered as well, indeed if the object has an appearance very similar to what is around, it is very likely to mistake it as part of the environment.
\item \textbf{Intra-class variation.} Categories often can have a low level of granularity, meaning that objects inside them could have substantial differences between each other, such that they could belong to different subcategories. A good model should take into account these diversities, understanding that these objects are part of the same category, even if they seem quite different.
\end{itemize}
As just pointed out many variables must be considered to have the image classification process working fine. In particular, a very good model must be the most invariant possible to all these aspects and, at the same time, maintain a good sensitivity to the inter-class differences.

\bigskip
\begin{figure}[ht]
 \centering
 \captionsetup{justification=centering}
 \includegraphics[scale=0.8]{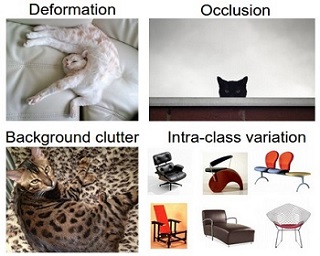}
 \caption{Visual challenges: deformation, occlusion, \\background clutter, intra-class variation}
\end{figure}
\bigskip

Thus, to recap, the described task consists in taking an array of pixels, representing an image, and assigning a label to it. To achieve the goal, a proper pipeline can be defined by the following steps:
\begin{itemize}
\item \textbf{Input.} A big dataset containing hundreds of thousands of annotated images, that are pictures coming together with the proper category they belong to. It is referred to be the training set given as input to a neural network, whose task is to exploit this bunch of images to understand what objects of interest look like. This approach could be called data-driven, as it relies on a training set made of labeled images.
\item \textbf{Learning.} The neural network relies on the data given as input and on the category associated to them, to perform a training. The learning approach is very similar to that which would be taken with a child: provide the network with many examples of each object-class, so to look at them and learn about their visual appearance. The output is represented by a model, a sort of visual knowledge about objects, that can be used to classify images.
\item \textbf{Evaluation.} The last step is to evaluate how accurate the model is. Its precision is evaluated by making it predict labels for a bunch of images, referring to known objects, it has never seen before.
\end{itemize}

\section{Object Segmentation}
Image segmentation \cite{ZAITOUN2015797} is another hot topic of computer vision. It is defined as the process of partitioning an image into multiple segments, so to turn its representation into something easier and more meaningful to analyze. The result of this process is a set of segments that all together cover the entire image, or a set of contours extracted from the image. The latter is the kind of result in which we are particularly interested, that is the process of detecting and extracting edges related to the object in the image. In the process of image interpretation or object recognition, objects shape provides much more information rather than their low-level appearance, but is hard to extract and represent, especially in situations where the object could be occluded or blended with the surrounding environment. Indeed in the section describing the setup of the experiment, several hypothesis will be made because of this issue. \\As mentioned at the beginning, we are interested in the process of identifying the whole boundary of an object, so to extract its shape and focus the classification task on this part only, leaving out all the other useless pixels related to the background.

\smallskip
\bigskip
\begin{figure}[ht]
 \centering
 \includegraphics{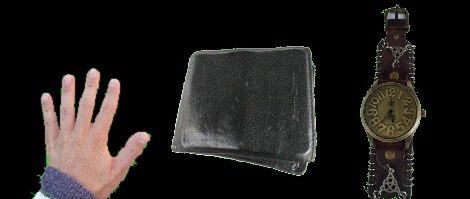}
 \caption{Examples of segmented objects}
\end{figure}
\bigskip
\smallskip

There are segmentation methods that are said to be trainable. In fact they use a Neural Network for segmentation purposes, similarly to what we do for classification. Using one of these segmentation techniques would require a huge domain-knowledge database to train a Neural Network, doing something very similar to what we already need for the classification step. For this reasons, this family of methods doesn't suite to our purpose, in fact we use a more traditional approach. In particular it relies on the appearance of the image itself, exploiting pixel-level information to smoothly define the region where the object is present. 
Each pixel belonging to a certain region shares similar characteristics or properties, such as color, intensity, or texture. On the other hand, adjacent regions are quite different from each other, considering the same characteristics. The algorithm we use takes advantage of the invariant features related to pixels that are part of the object, to differentiate them from those composing the background and to extract its boundary.

\section{Object Discovery}
In this work object discovery is defined as the problem of understanding where an object is in a scene. We, as humans, haven't any problem entering an unknown place, look for an object we need and instantaneously recognize it. It happens so fast that we don't even realize the amazing work done by our visual recognition system. Thinking about the same situation for a robot, it is quite evident that the whole process becomes much more complex. That's why we are interested in addressing a typical robotics problem, where, for instance, a robot enters a human place and has to find an object that it was asked about.
We imagine that the robot starts looking around, capturing video scans of the surrounding environment, trying to "see" the object it is looking for. The idea is to tackle this recognition problem by extending the classification task. In detail, from every scan a sequence of images could be extracted, according to a proper frame rate, so to capture all the relevant information. The trade-off is to choose a frame rate that optimally fits the scanning speed of the robot: too few frames would result in missing visual information, while too many would cause a significant increasing in processing time. Every frame is then processed with the Neural Network model used for single classification, so to finally understand the location of the object.
\bigskip
\section{Database Creation}
The main limitation of databases containing annotated images is that even the largest one counts a finite number of elements, as the collection of labeled picture is really time-expensive. In addition to this, they are quite general-purpose, as they include a large variety of object categories meant to satisfy all the needs. The latter is the reason why often one would like to have a more specific training set that better fits its tasks. That's why we propose a way to quickly gather an object appearance from different views. The idea is to have a tool that allows to take a scan of an object, considering a range that captures all its angles, and automatically saves its main views, together with the proper label, inside a database. The scan is meant as a short video, during which, for instance, the object is rotated around the vertical axis making use of a turning table. The video is then broken up into frames which are selected in such a way to capture substantial variations in object orientation.
\chapter{Related Works}
We now list the main works that dealt with problems related to ours, dedicating a different section to each topic.
\section{Database Creation}
Creating databases containing high-quality annotated images plays an important role in having intelligent systems able to extract knowledge from visual information. A very famous and common approach is to use a turntable, placing an object onto it and acquiring photograms while rotating the table around its vertical axis. Several works rely on this method to capture objects appearance, such as COIL-100 \cite{nayar1996columbia} and COIL-20 \cite{nene1996columbia} where they build a database of object images using a motorized turntable and a black background, with the former considering colored images of a hundred object classes and the latter focusing on gray-scale ones of 20 classes. In both cases the table is rotated through 360 degrees, while taking object images at intervals of 5 degrees to vary object pose with respect to the camera. Gathering visual data using a turning table can have several applications and can be performed in many different ways. For instance \cite{reinhold2001image} proposes an image database for 3-D object recognition made of objects taken from the office and health care domain. It is a 13-categories dataset built by placing such objects on a turntable and taking images with a camera mounted on a robot, so that it could be lifted
and lowered. In such scenario the image taking process considers a dark background and more than 3000 viewpoints uniformly distributed on a hemisphere-like surface above the object, so to faithfully capture its angular view variations. Another example of how useful this approach can be is presented in \cite{Borji2016iLab20MAL}, where they overcome the problem that most of large-scale databases are not very useful to learn network invariance to parameters such as translation, pose, illumination and background, introducing a large-scale turntable-based dataset and using it to deeply inspect CNNs invariance properties. Anyway what is proposed in our work is somehow different, as the portability and mobility of smart phones give access to an incomparable freedom and possibility of collecting really task-specific databases.
\section{Object Discovery}
Entering a certain environment and recognize previously seen objects is a task that we, as humans, accomplish quite simply relying on our past visual knowledge \cite{kemp2009object}. Differently, a robot in the same situation has quite trouble in finding the object that is looking for, reason why several attempts have been made to develop deep learning methods capable of recognizing objects appearance from scene shortcuts. Such a task requires the construction of good features, where good means being invariant to non relevant input variations while at the same time preserving significant information.
Deep Convolutional Networks constitute trainable architectures that can learn such invariant features from large amounts of labeled training samples \cite{lecun2010convolutional}. The authors of \cite{ciregan2012multi} use winner-take-all neurons, determining them with a max pooling operation, to train DNN columns and average their prediction for recognizing objects, while \cite{cirecsan2011high} presents a fast trainable CNN implementation for handwritten digits recognition and image classification. DCNNs have been established as a powerful class of models for visual recognition problems and can be used also on large-scale video classification \cite{karpathy2014large}. Our work relies on training a very famous, successful deep architecture on a large-scale dataset of annotated images and wrapping it inside a mobile-based system to perform object recognition. In particular a short video of a scene is analyzed by the system and classified according to the most likely object between the top-5 being detected.
\newpage
\section{Deep learning on mobile platforms}
Deep learning is starting to have success on mobile platforms too. We are moving towards intelligent and sophisticated user applications, related to different fields, that may use deep learning techniques to provide unique functionalities to users. Keeping in mind that a picture is worth a thousand keywords, an important example is represented by image-based object search for mobile platforms \cite{neven2009image}, \cite{yeh2005picture}. Such applications allow to retrieve image-based information from the web by taking the object appearance as input and connecting to a server to find web pages whose images match with the selected one. The server has an object recognition engine for assigning a confidence score to its visual knowledge based on the image received from the phone and then generating an output in line with the requested query. Instead \cite{bruns2008mobile} proposes a project to overcome the problem of museum audio guides, that are low informative and not very intuitive, by enabling conventional smart phone camera with museum guidance abilities. In this way museum visitors would have access to easier multimedia functionalities, such as automatic object recognition by taking non-persistent photos, that would also save museum owners from the cost of devices, as every visitor would be able to have its own phone equipped with such tools. Visual learning can also be applied to security, such as recognizing the iris \cite{Cho2006PupilAI} for managing a bank transaction with mobile phones. Anyway we didn't find any work whose goal is to integrate deep learning inside the mobile platform for robotics purposes. Our system is thought to abstract from the need of a physical robot, as having one could be complicated and expensive, to have a highly-portable, low-cost tool for testing visual models in real robotics situations.
\newpage
\chapter{System Components}
In this chapter we present the relevant components of the whole system, paying particular attention to the classification and segmentation modules. 
As a general explanation of its functioning, the user interacts with a welcome interface, where it can choose one of the visual tasks wrapped inside the app between a list of proper buttons. According to the choice, one of the two main activity branches is expanded: if the user chooses classification with or without segmentation, the option to take a picture or select one from the gallery is shown and then the proper activity related to the camera or to the file selection is started. Differently if the choice is about object discovery or database creation, a call to the camera is performed in order to open it in video mode, as the user needs to register a scan of the environment. Then it is checked which of the two options was selected to properly process the video scan. In particular, for database creation some frames are extracted from the video, in order to catch the object of interest under different views, and saved into the gallery, while object discovery performs classification of video snapshots and returns the most likely frame with the object inside. On the other hand, the classification branch checks if the segmentation option was selected and, in that case, segments the object out of the image. Finally the figure, segmented or not, is classified and returned together with the label of the most likely object. The app apk can be found at \textbf{https://drive.google.com/open?id=0B8JAK5RIeaD9Ni01TkNKSEs3Y00} together with instructions on how to make it work. In the following we show a flow scheme to quickly understand the structure of the app and its main building blocks.

\begin{figure}[p]
 \centering
 \includegraphics[scale=0.85]{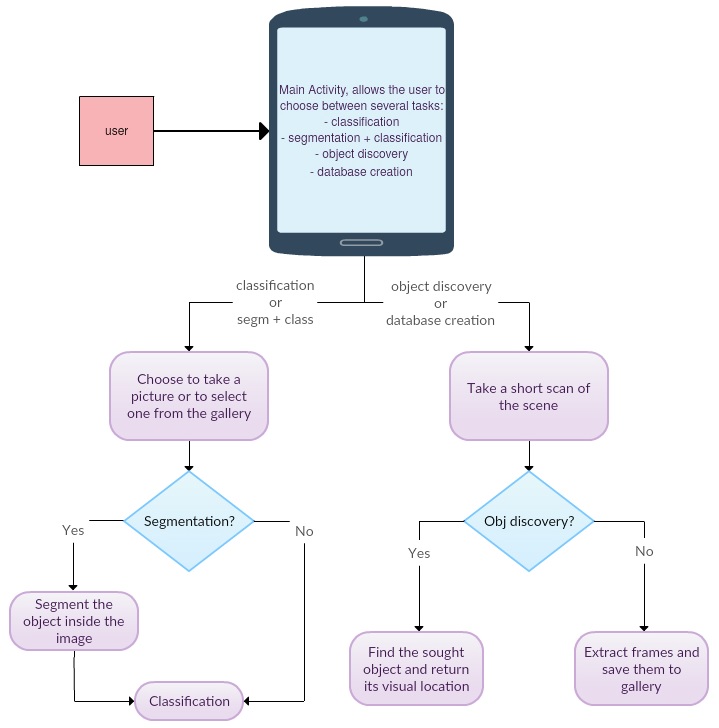}
 \caption{Application flow chart: first is the main graphical user interface, left branch refers to operations and computations needed to perform classification or segmentation tasks, right branch deals with video manipulation for object discovery and database creation}
\end{figure}
\bigskip

\newpage
\section{Classification module}
Now we are going to have a deeper insight into the classification module. In particular we present the framework used for training, showing how input and output of a layer are managed and how to specify the different layers that a network is made of. We describe also the network architecture, focusing on the nature of layers and the used activation function, as well as the dataset used for training. 
\subsection{Caffe}
Caffe's birth has been highly motivated by large-scale visual recognition and in particular by the excellent results achieved in this field by the so called Convolutional Neural Networks. The arising problem was that replying results obtained with these architectures could involve months of work by researchers because of the missing of computationally efficient toolboxes for the deployment of state-of-the-art models.
Caffe \cite{jia2014caffe} came as an open-source framework for deep learning that allows to access deep architectures and use them in a very straightforward way.

\bigskip
\begin{figure}[ht]
 \centering
 \includegraphics[scale=0.53]{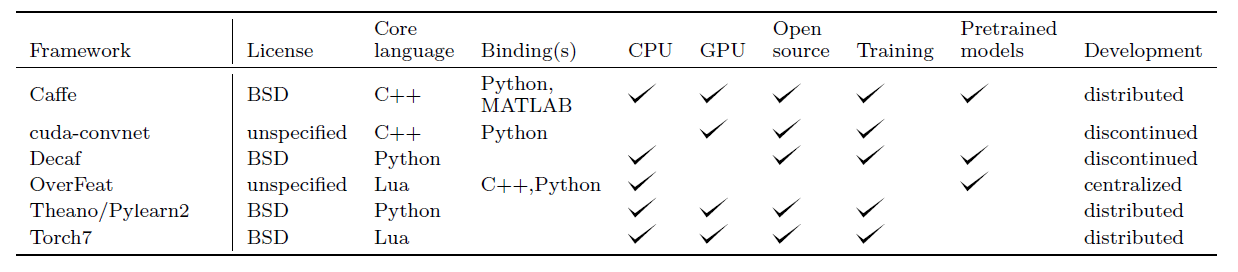}
 \caption{Comparison of popular deep learning frameworks}
\end{figure}
\bigskip

In this thesis we use Caffe as it provides a complete toolkit for training and testing models, with a good documentation about what one needs to install to use the framework and instructive examples for all the tasks. It also includes a whole set of layer types, such as: convolution, pooling, inner products, element-wise operations, besides losses like softmax and hinge.\\
A Caffe layer manages data in the form of blobs, taking one or more of them as input and providing one or more of them as output. A blob is an envelope put around the actual data and is mathematically defined as an N-dimensional array, with N usually counting 4 dimensions. 
These structures are used by Caffe layers to store and communicate data, such as: batches of images, model parameters, derivatives for optimization.\\
A layer is the gist of a network model, as its fundamental unit of computation, and it is initialized once for all, together with its connections, at model initialization. Beyond the setup, each layer is responsible for two critical operations: a forward pass and a backward pass.

\bigskip
\begin{figure}[ht]
 \centering
 \includegraphics[scale=0.53]{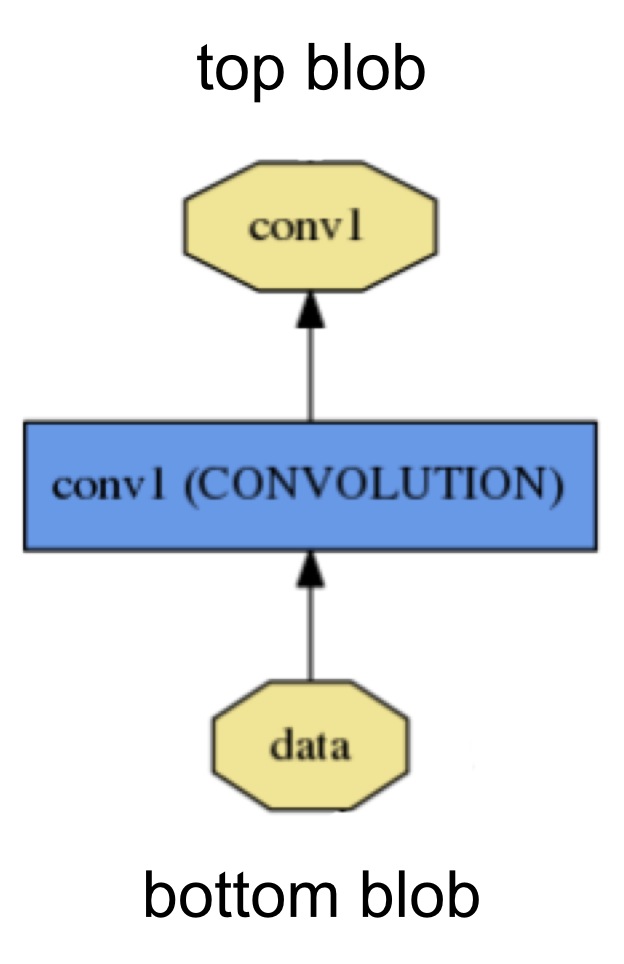}
 \caption{Example of layer input and output}
\end{figure}
\bigskip

Forward and backward functions are implemented both for CPU and for GPU, allowing to choose the best option according to the kind of experiment one needs to do.
The forward takes an input from bottom, computes the output and sends it to the top. While the backward computes the gradients with respect to the parameters and the inputs and back-propagates them to earlier layers. Forward and backward are iteratively called by the Solver, an important Caffe module, that takes care of optimizing the model incorporating its gradient into a weight update for achieving loss minimization.
In particular, at each iteration, the Solver makes a forward call to compute output and loss and a backward call to compute the gradients, then includes the latter into parameter updates and changes its state according to learning rate and history.\\
The set of all layers together with their connections define the whole network, that is specified in a plaintext protobuf schema.

\bigskip
\begin{figure}[ht]
 \centering
 \includegraphics[scale=0.8]{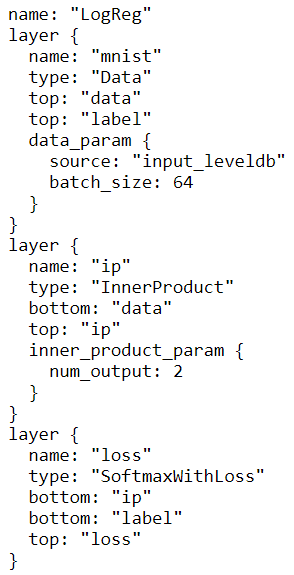}
 \caption{Example of logistic regression classifier specified in protocol buffer schema}
\end{figure}
\bigskip

From the example one can quickly understand that this network is made of three layers: the first accounting for the input data, the last being the loss layer that computes the objective for a specific task and one hidden layer that deals with inner product operation.
This modeling language is used as it is self-explanatory, human-readable and easily implementable with multiple languages, all properties that give to it incomparable flexibility and extensibility.
\newpage
\subsection{Network Architecture}
Objects in real situations show significant variability, thus in order to learn to recognize them it is necessary to consider very large training sets, that's why we use the ILSVRC 2012 \cite{ILSVRC15} dataset for this purpose. Anyway to learn about a thousand objects from more than one million images, a model with a great learning capacity is needed.
Using regular Neural Nets wouldn't be possible as they are characterized by a fully-connected structure, where each neuron of a certain layer is connected with all the neurons of the previous layer. For instance consider a situation where images of size $32\times32\times3$ (width, height, number of channels) are taken as input, in here one neuron in the first hidden layer would have $32\cdot32\cdot3 = 3072$ weights, that is already a significant amount even if we are referring to quite small images. This example suggests that the regular structure wouldn't scale well to larger images. In fact, considering an image of more usual size, e.g. $200\times200\times3$, the number of weights increases vertiginously to $200\cdot200\cdot3 = 120,000$ weights. For this reason we rely on Deep Convolutional Neural Networks (DCNNs), an evolution of usual Neural Networks whose main difference is to take advantage of the input nature. A DCNN exploits the fact that input is constituted by images to properly constrain its architecture, indeed its neurons are represented by 3D volumes counting three dimensions: width, height and depth. 
Neurons in a layer are only focused on a small image region, instead of considering all of it as in fully-connected networks, overcoming in this way the problem of scaling to larger images.

\bigskip
\begin{figure}[ht]
 \centering
 \includegraphics[scale=0.65]{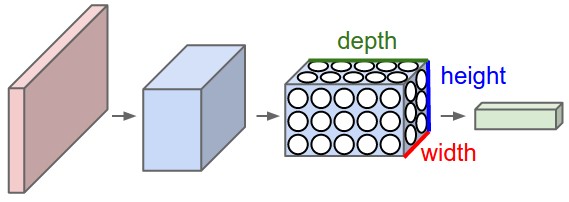}
 \caption{Neuron's structure inside a ConvNet}
\end{figure}
\bigskip

For our training we consider a particular architecture: the Alexnet \cite{NIPS2012_4824} Deep Convolutional Neural Network. Its structure is made of eight layers, five of them convolutional and the remaining three fully-connected:\\

\bigskip
\begin{figure}[ht]
 \centering
 \includegraphics[scale=0.65]{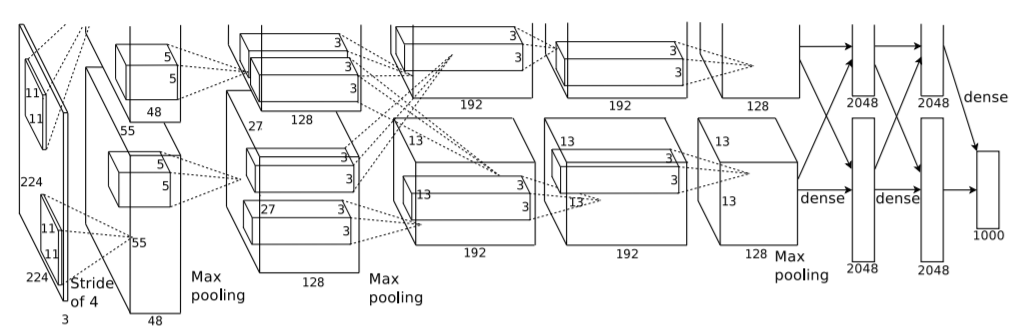}
 \caption{AlexNet architecture}
\end{figure}
\bigskip

\begin{itemize}
\item \textbf{Convolutional.} The input is represented by the raw pixel values of the image, in this case having dimension $224\times224\times3$. First layer is convolutional, the layer's type that is the core of a Convolutional Network and has a set of learnable filters as parameters. 
Essentially each filter refers to a neuron connecting to only a local region of the input volume, instead of establishing connections with all neurons in the previous volume. The spatial region covered by the connection is a hyperparameter called receptive field of the neuron. In our case the receptive field, or filter size, has dimension $11\times11\times3$ and refers to a sub-area of the whole image as it covers its full depth. Convolution consists in sliding each filter along the two main image dimensions, according to a fixed stride that determines how many pixels at a time are moving, while computing dot products between its entries and those of the input image. For each filter, this procedure yields a 2-dimensional activation map that keeps information about the different replies that the filter gave in the visited spatial regions. The idea is that the network will learn filters activating in correspondence of some visual feature, considering simpler ones for the first layer and more and more complex as the layers go higher. The output volume of the layer is made by stacking up the activation maps, along the depth dimension, related to all the filters. Considering a stride of 4 and a number of filters equal to 96, here the output is represented by a block having dimension $55\times55\times96$ that in the image is split in case one uses two GPUs for training. The rest of layers up to the fifth are all convolutional and follow similar rules, only with different values. 
\item \textbf{Max Pooling.}
Successive convolutional layers often could have a Pooling layer between them. Its main objective is to reduce the spatial size of the representation, so to decrease the number of parameters and amount of computation in the network. This layer separately takes every depth slice of the input volume and applies to it the MAX operation to reduce its size. It takes the maximum over a certain number of activations, defined by the dimension of its filters, discarding the rest, cutting in this way width and height of the input, but leaving its depth unchanged. In the AlexNet architecture a variant of traditional pooling is used, referred to be the overlapping pooling. Considering a filter size of $z\times z$, the main difference is represented by the stride that in this case is lower than $z$, causing in this way an overlap between the windows performing the pooling. In particular, with this architecture, a stride of 2 and a $z$ equal to $3$ are used, as they lead to a slight improvement in the error rate with respect to the non-overlapping scheme.
\item \textbf{Fully-connected.} 
The last three layers are characterized by a fully-connected structure of 4096 neurons, where each neuron is connected to all activations in the previous layer. The output of the last layer is given as input to a softmax operating over a thousand categories, as the wanted result is a distribution of probabilities indicating which label is more likely to be the correct one.
In particular, the softmax function is defined as $f_j(z)=\frac{e^{z_j}}{\sum_{k} e^{z_k}}$, takes as input a vector of real-valued scores and maps it to a vector taking values between 0 and 1, referring to the sought probability distribution.
\end{itemize}
\newpage
The neurons output is modeled according to a function of their input x, that is $f(x) = max(0; x)$. Neurons with this kind of nonlinearity are referred to be Rectified Linear Units (ReLUs).

\bigskip
\begin{figure}[ht]
 \centering
 \includegraphics[scale=0.65]{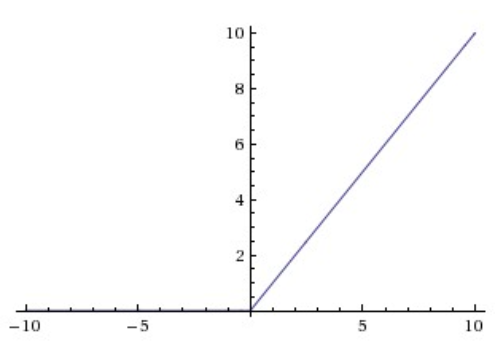}
 \caption{Rectified Linear Unit (ReLU) activation function}
\end{figure}
\bigskip

In terms of stochastic gradient descent convergence time, they allow a DCNN training that is several times faster than equivalents with saturating, usual neurons, due to the non-saturating nature of this nonlinearity. Moreover, with respect to other ways of modeling the output of a neuron, such as $f(x)=tanh(x)$ or $f(x)=(1+e^{-x})^{-1}$, that involve expensive operations, the ReLU can be simply implemented by imposing a zero threshold on activations matrix.\\
Anyway there's a delicate aspect about its fragility during training, that one has to deal with when using ReLU. Indeed, it can happen that, due to a big gradient passing through the activation and the consequent updating of weights, the neuron won't activate on any datapoint anymore. In particular, the gradient passing through the unit, starting from this point, will always be zero. Basically ReLU units can stop working during training, without activating across the entire dataset anymore. If a too high learning rate is chosen, this problem could really cause a great percentage of networks to die. Anyway with a proper tuning of the learning rate, the mentioned issue happens less frequently and as a matter of fact ReLU is the simplest and most used activation function for DCNNs.
\newpage
\subsection{Training Set}
The huge set of labeled pictures used to train the AlexNet architecture for classification is part of a bigger one that's called ImageNet \cite{5206848}. The latter is a very big image dataset organized according to WordNet \cite{Miller:1995:WLD:219717.219748} hierarchy. Basically ImageNet provides, more or less, a thousand images for most of all synsets, where a synset is a meaningful WordNet concept. Moreover, using WordNet as a backbone, care is taken about words with the same spelling, but referring to different classes, or being diverse but related to the same object, such as synonyms. The labels attached to every image, that summarize their content, have been provided by humans, creating in this way the largest existing set of annotated images that also satisfies a high level of pictures quality.
ImageNet contains more than 15 million labeled images approximately spread over 22,000
categories. The dataset we use as training set is called ImageNet Large-Scale Visual Recognition Challenge 2012 (ILSVRC '12) and can be downloaded directly from the official site: \textbf{www.image-net.org/challenges/LSVRC/2012}. It is a slice of the whole collection and has become a standard benchmark for large-scale visual recognition. The process of constructing an object recognition dataset of this dimension has a main step, that is the definition of the set of target object categories. In particular they are selected from the existing ImageNet categories,  combining automatic heuristics and manual post-processing
to create the list of target classes for each task. Its annotations can belong to two main types: image-level annotation of a binary label, pointing out the presence
or absence of a certain object in the image, and object-level annotation giving information about the class label of the object instance in the image, with this latter type being of main interest for our classification task. ILSVRC contains more than one million annotated images spread over a thousand object categories and comes from an annual competition, started in 2010, that has the same name.
The scale of this challenge dataset allows to have unprecedented opportunities in evaluating and developing object recognition algorithms, creating, for instance, the need for techniques able to distinguish objects that are visually very similar, reason why it has become a standard in training Neural Networks for visual tasks. 
\newpage
\section{Segmentation Module}
Here we present the structure of the segmentation module, describing the relevant blocks of its algorithm, but leaving more technical details for the next chapter.
First of all it must be said that the app, in order to apply the segmentation algorithm to the image, connects to the server where such algorithm is hosted. This is not a design choice, as at the beginning the whole module was planned to reside locally, but more a forced solution due to a great limit, found in implementation phase, that we will explain in more details later on. The architecture is similar to that of Software as a Service in Cloud Computing Technologies. Application internal data cannot be accessed, thus integration protocols and application programming interfaces (APIs) are used to operate over a wide area network.  

\bigskip
\begin{figure}[ht]
 \centering
 \includegraphics[scale=0.7]{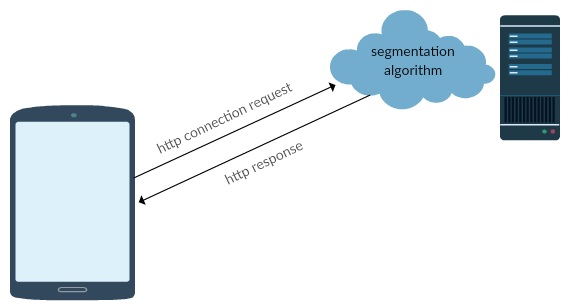}
 \caption{Segmentation module connecting to server to segment the image}
\end{figure}
\bigskip

If the classification module positively checks that the user chose to segment the image, before classifying it, then the focus is passed to the segmentation part. In here data are prepared to be sent to the server through a channel established by an HTTP connection. Once the image arrives on the server, it is processed, segmented and then sent back to the app for further computations.\\
About the segmentation algorithm structure, it is not based on any deep learning method. It's more a geometrical approach to develop a light solution, without the needing of a network training, to help the work done by classification. The method exploits image appearance and its geometrical properties to pull the object out of it for simplifying further processing.
\newpage
The whole process can be identified by four main blocks, one after each other, as follows: 

\bigskip
\begin{figure}[ht]
 \centering
 \includegraphics[scale=0.7]{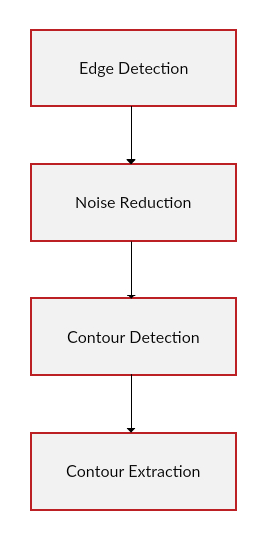}
 \caption{Block diagram of the segmentation algorithm}
\end{figure}
\bigskip

The first submodule takes as input an image coming from the app and performs edge detection by computing differences between pixels intensities along its height and width. Of course, since we deal with color images, the whole process must be run on every RGB channel, detecting edges for each color and then combining them to find the final image edges. Once we have the edged image, we feed the Noise Reduction block with it. This one concerns about reducing image noise by collapsing any pixel intensity lower than the mean of all edged image intensities. Then the image is passed to Contour Detection that is in charge of detecting contours in a hierarchical way. In particular any contour that is inside another one is treated as a child of the latter. In this way contours can be nested to more than one level, resulting in a tree-like structure. Finally, in order to only extract the contour relevant to the object, we delete any contour that doesn't cover at least a certain percentage of the whole image and return the remaining one as main boundary of the object of interest.
\chapter{System Implementation}
This chapter describes the implementation choices that have been made to realize our system. In here we present and describe the adopted programming environment, as well as the main libraries that have been used. We prefer to divide the body of the chapter in different parts, each one describing technical and implementation details regarding a dedicate section of the system.
\section{Classification Module}
In this section we give insight about the Java Native Interface (JNI), which allows us to import native libraries for Caffe and use their methods to deal with deep learning actions in a Java Environment. We also describe the whole block that deals with the accomplishment of the classification task from a Java point of view.
\subsection{JNI and libcaffe}
The Java Native Interface can be considered as a Java framework that allows Java code to call a piece of native code, where the adjective native means that it is written using another programming language, usually C/C++. The main goal for which JNI has been developed is the possibility to call code pieces in charge of some particular task from within Java programs. The reason why this tool is so useful is that it gives the opportunity of using code that implements some particular, non-portable functionality that can't be written in pure Java from scratch. The interface is based on defining several class, having a Java interface, that are in charge of performing the mapping between the two environments, but leave the implementation of their methods to the native code. The interface can define an arbitrary number of native methods, that have the distinguishing characteristic of being declared with the usage of keyword "native" and have an empty body, as they are already implemented in the related native library. The access to this kind of methods happens in a rather indirect way, as the real native functions are called from those functions implementing native methods that are in turn used by Java code. In our work, we have two couples of native code libraries, that are \textbf{libcaffe.so} and \textbf{libcaffe\_jni.so}, each one dealing with the support for a different family of processors. The former gives support to the \textbf{arm64-v8a} family, while the latter for \textbf{x86\_64}, that are both 64-bit architectures for Android platforms, but supporting different instruction sets. The class playing the role of glue between the two environments is \textbf{CaffeMobile.java}. It mainly defines native methods included in the mentioned libraries and some extensions of them useful to accomplish the different visual tasks. In the following the implementation of such a class is shown, highlighting its most relevant native methods without access modifiers, such as public and private, for the sake of a good visualization:
\bigskip
\begin{lstlisting}
public class CaffeMobile {
    static byte[] stringToBytes(String s) {
        return s.getBytes(StandardCharsets.US_ASCII);
    }
\end{lstlisting}
\bigskip
\textit{stringToBytes(String s)} is a static method as it doesn't need any object instance for being called and belongs to the class itself, without depending on the individual characteristics of the class. It takes as input a Java string and return its conversion to an ASCII byte array. It is used by Java methods of CaffeMobile class to reconstruct bytes of data associated to an image starting from its path, to pass it as parameter to native functions.
\bigskip
\begin{lstlisting}
native void setNumThreads(int numThreads);
\end{lstlisting}
\bigskip
\textit{setNumThreads(int numThreads)} is one of the methods coming from native libraries. It is in charge of changing the number of cores that we use to process an image, as using a higher number of threads leads to better effective performance.
\bigskip
\begin{lstlisting}
native int loadModel(String modelPath, String weightsPath);
\end{lstlisting}
\bigskip
This native method is one of the most important as its role is to load the model to be used for image classification. It takes as input a couple of strings, the former referring to the file path related to \textit{"deploy.prototxt"} inside which the structure of network layers is specified and the latter referring to the path of the caffe model, that consists in the learned weights resulting from the training procedure.
\bigskip
\begin{lstlisting}
void setMean(float[] meanValues) {
        setMeanWithMeanValues(meanValues);
}
    
native void setMeanWithMeanValues(float[] meanValues);
\end{lstlisting}
\bigskip
\textit{setMean(float[] meanValues)} is an example of an indirect call to a native method. Indeed it is a Java method that takes as input a float array containing image mean values computed on the whole training set and inside its body calls the real native method responsible for the update of such values.
\bigskip
\begin{lstlisting}
float[] getConfidenceScore(String imgPath) {
   return getConfidenceScore(stringToBytes(imgPath), 0, 0);
}
    
native float[] getConfidenceScore(byte[] d, int w, int h);
\end{lstlisting}
\bigskip
\textit{getConfidenceScore(String imgPath)} takes as input the path of the image that we are classifying, reconstructs data bytes associated to it by using \textit{stringToBytes} defined before and pass them, together with image width and height, to the proper native method in charge of computing the corresponding accuracy scores for all the object categories. The output is an array of floats containing a thousand confidence values, each one representing the probability that its related label is the correct one.
\bigskip
\begin{lstlisting}
int[] predictImage(String imgPath) {
    return predictImage(imgPath, 1);
}

int[] predictImage(String imgPath, int k) {
    return predictImage(stringToBytes(imgPath), 0, 0, k);
}

native int[] predictImage(byte[] d, int w, int h, int k); 
\end{lstlisting}
\bigskip
\textit{predictImage(String imgPath)} is the Java method that manages the classification process from within the MainActivity. In here the call to the native method is even more nested, as the first one takes as input the image path and pass it to a second Java method, together with a value, that by default is 1, expressing how many labels one wants back as classification output. The latter calls the native method, passing image data, dimensions and number of labels as parameters and gets back an array of integer values being the output labels for the selected image, sorted in order of accuracy.

\subsection{Single Image Classification}
The first task that is implemented is a simple single image classification. The user can access to it by simply pressing the proper button from the welcome interface. As soon as there's a request of such a type, the Java \textit{classification()} method starts:
\bigskip
\begin{lstlisting}
public void classification () {
 setContentView(R.layout.classification);

 ivCaptured = (ImageView) findViewById(R.id.ivCaptured);
 tvLabel = (TextView) findViewById(R.id.tvLabel);

 btnCamera = (Button) findViewById(R.id.btnCamera);
 btnCamera.setOnClickListener(new Button.OnClickListener() {
  public void onClick(View v) {
    initPrediction();
    fileUri = getOutputMediaFileUri(MEDIA_TYPE_IMAGE);
    Intent i = new Intent(MediaStore.ACTION_IMAGE_CAPTURE);
    i.putExtra(MediaStore.EXTRA_OUTPUT, fileUri);
    startActivityForResult(i, REQUEST_IMAGE_CAPTURE);
    }
  });
\end{lstlisting}
\bigskip
The user is shown with another screen where it can choose between capturing the image to be classified from the camera or selecting it from those saved in the gallery. For both actions a listener is defined to check for all the events related to its button. In the former case, when the camera button is pressed, a method that activates on that click only starts. The relevant part is about recovering the Uniform Resource Identifier (URI), that is basically a string of characters used to identify a resource, and attaching it to an Intent together with a Standard Intent action, \textit{ACTION\_IMAGE\_CAPTURE}, necessary to have the camera application capture an image and return it. Such an Intent and a proper request code related to the particular kind of action are used as parameters to call \textit{startActivityForResult()}, a method that deals with the execution of a specific activity and returns a \textit{RESULT\_OK}, if everything went well, and an Intent carrying the resulting data. In the latter case, that is about selecting a saved image from the gallery, the code is very similar, but changes relevant parameters accordingly, to take into account the diverse type of action, ending with a call to start the activity and wait for its results. The following piece of code completes the \textit{classification()} method introduced previously.
\bigskip
\begin{lstlisting}
btnSelect = (Button) findViewById(R.id.btnSelect);
btnSelect.setOnClickListener(new Button.OnClickListener() {
  public void onClick(View v) {
    initPrediction();
    Intent i = new Intent(Intent.ACTION_PICK,
    MediaStore.Images.Media.EXTERNAL_CONTENT_URI);
    startActivityForResult(i, REQUEST_IMAGE_SELECT);
    }
  });
}
\end{lstlisting}
\bigskip
When the user ends the chosen action, the system calls \textit{onActivityResult()} method on the activity that was selected. It takes as input three arguments: the request code passed to \textit{startActivityForResult()}, a result code returned by the activity, that can be \textit{RESULT\_OK} if it was successful or \textit{RESULT\_CANCELED} if it failed, and an Intent that carries the resulting data. This method is in charge of handling all the returning activity types, distinguishing between them in its body, according to the returned parameters.
\bigskip
\begin{lstlisting}
void onActivityResult(int reqCode, int resCode, Intent d) {
if ((reqCode == REQUEST_IMAGE_CAPTURE || reqCode ==
REQUEST_IMAGE_SELECT) && resCode == RESULT_OK) {

if (reqCode == REQUEST_IMAGE_CAPTURE) {
  imgPath = fileUri.getPath();
}

else {
  Uri selectedImage = d.getData();
  String[] filePathColumn = {MediaStore.Images.Media.DATA};
  Cursor cursor = MainActivity.this.getContentResolver().
  query(selectedImage, filePathColumn, null, null, null);
  cursor.moveToFirst();
  int columnIndex = cursor.getColumnIndex(filePathColumn[0]);
  imgPath = cursor.getString(columnIndex);
  cursor.close();
}
\end{lstlisting}
\bigskip
The request code is handled to understand which of the two actions was performed and then save the resource path. In case the image was captured, its path is simply retrieved from the fileUri, as a direct consequence of \textit{ACTION\_IMAGE\_CAPTURE}. On the other hand, retrieving the path of a selected image is a bit longer. A Cursor object is defined with the image URI, retrieved from the Intent, and the columns to return obtained from \textit{MediaStore.Images.Media.DATA}. Then the Cursor is positioned on the first row and the image path is retrieved with \textit{cursor.getString(columnIndex)}, where \textit{columnIndex} is the first column name between those returned. Once we obtain the path, we are able to think about classifying the image.\\
To classify the image it's necessary to invoke the proper native methods that will be described in a moment. Anyway it is quite a heavy task, thus in order to not stress too much the main thread, all the operations are wrapped inside an asynchronous task that runs in background, while a progress dialog is shown to let the user know that the process is running. Such a task is defined as \textit{cnnTask} and executed, passing as input the path of image to be predicted.
\bigskip
\begin{lstlisting}
dialog = ProgressDialog.show(MainActivity.this,
"Predicting...", "Wait for one sec...", true);
CNNTask cnnTask = new CNNTask(MainActivity.this);
cnnTask.execute(imgPath);
\end{lstlisting}
\bigskip
\textit{CNNTask} is a private class extending \textit{AsyncTask}. The latter is an abstract class that allows to perform background operations and publish results on the User Interface thread. \textit{CNNTask} overrides some of its main methods, such as \textit{doInBackground()} and \textit{onPostExecute()}, to accomplish classification. This class defines and initializes also a \textit{CNNListener} object, where \textit{CNNListener} is a simple interface, implemented by the Main Activity, whose body declares only \textit{onTaskCompleted()} method. The latter is called inside the asynchronous task, as soon as the background operations are completed, to show the final result, anyway its implementation and role's explanation will follow later on.
\bigskip
\begin{lstlisting}
class CNNTask extends AsyncTask<String, Void, Integer> {
  private CNNListener listener;
  public CNNTask(CNNListener listener) {
  this.listener = listener;
}
  @Override
  protected Integer doInBackground(String... strings) {
    return caffeMobile.predictImage(strings[0])[0];
  }
\end{lstlisting}
\bigskip
\textit{AsyncTask} uses three types of parameters \textit{<Params, Progress, Result>} , where \textit{Params} represents the type of parameters sent to the task upon execution, \textit{Progress} refers to the type of progress units published during the background computation and \textit{Result} accounts for the type of result of the background computation. \\The method \textit{doInBackground()} is in charge of performing the fundamental computations on a background thread. The specified parameters are those passed to \textit{execute()} by \textit{cnnTask} caller, that in this case consists in an array of strings, with image path saved in the first position. In simple classification the body of this method is only constituted by a call to \textit{predictImage()}, that is part of \textit{caffeMobile} class described before, which in turn calls the native method to predict an image label and returns an array of integers representing all the labels sorted in order of confidence. Finally, the first array element, that in fact is the most likely label, is returned as result of \textit{doInBackground()}. As soon as that integer value comes back, it is passed as parameter to \textit{onPostExecute()} method that deals with performing the needed operations after that the main task is completed.
\bigskip
\begin{lstlisting}
  @Override
  protected void onPostExecute(Integer integer) {
    listener.onTaskCompleted(integer);
    super.onPostExecute(integer);
  }
\end{lstlisting}
\bigskip
The method essentially takes the resulting label as input and delegate to the \textit{onTaskCompleted()} method the job of ending the whole process. 
\bigskip
\begin{lstlisting}
  @Override
  public void onTaskCompleted(int result) {
    bmp = BitmapFactory.decodeFile(imgPath);
    ivCaptured.setImageBitmap(bmp);
    tvLabel.setText(IMAGENET_CLASSES[result]);
  }
\end{lstlisting}
\bigskip
To make result visible on the smart phone, \textit{onTaskCompleted()} uses \textit{decodeFile()} method of the Android class \textit{BitmapFactory} to decode image file path into a Bitmap object. Such a Bitmap is then set as the content of the ImageView \textit{ivCaptured}, while the TextView \textit{tvLabel} is updated with the proper image annotation. In particular the integer value returned from the previous task is used as index of \textit{IMAGENET\_CLASSES} array, that keeps all the thousand classes description, to retrieve the specific label and show it on the screen.\\
\smallskip
Remembering that this whole process just described takes place in background, we now complete the module explanation pointing out how the operations related to model and labels are handled.
\bigskip
\begin{lstlisting}
caffeMobile = new CaffeMobile();
caffeMobile.setNumThreads(4);
caffeMobile.loadModel(modelProto, modelBinary);
caffeMobile.setMean(meanValues);
AssetManager am = this.getAssets();
try {
  InputStream is = am.open("synset_words.txt");
  Scanner sc = new Scanner(is);
  List<String> lines = new ArrayList<String>();
  while (sc.hasNextLine()) {
  final String temp = sc.nextLine();
  lines.add(temp);
  }
IMAGENET_CLASSES = lines.toArray(new String[0]);
} catch (IOException e) {
e.printStackTrace();
}
\end{lstlisting}
\bigskip
\textit{caffeMobile} object is an instance of the \textit{CaffeMobile} Java class previously described, thus it can access all the fundamental methods that make the mapping between native code and Java possible. First it is set the number of threads, carefully considering the device architecture on which the system will run. Then the model obtained after training is loaded with the native \textit{loadModel()} method, passing as parameters the files containing respectively its structure and binary. The last action concerning native code deals with setting image mean values obtained from the dataset before training. While, regarding the labels, they are loaded through an Asset related to the text file containing them. Every label is stored in the \textit{IMAGENET\_CLASSES} array in a subsequent way, respecting the original order.
\subsection{Object Discovery}
A direct extension of single image classification is represented by the task of object discovery. The latter consists in finding a certain object, specified by an input label, through the analysis of a scene. In particular a scan, or short video, of an environment is taken, with the goal of recognizing a certain known object that the robot is searching. The code for this part is very similar to that related to single classification, until the camera is opened in video mode instead of image mode. The user plays the robot's role by taking a small video while slowly sliding horizontally, as to simulate the action of looking. As soon as the video capture action is completed, we execute an asynchronous task dealing with extracting video frames for further processing. Again this is a quite heavy operation, reason why we wrap it inside a background process, so to not strain too much the main thread.
\bigskip
\begin{lstlisting}
new extractFrames().execute();
class extractFrames extends AsyncTask<Void, Void, Void> {

@Override
protected Void doInBackground(Void... voids) {
MediaMetadataRetriever mediaMetadataRetriever = new
MediaMetadataRetriever();
mediaMetadataRetriever.setDataSource(videoPath);


for (int i = 0; i <5; i++){
Bitmap f = mediaMetadataRetriever.getFrameAtTime(i*1000000);
try {
  f.compress(Bitmap.CompressFormat.JPEG, 100, out);
} catch (Exception e) {
  e.printStackTrace();
}
}
return null;
}
\end{lstlisting}
\bigskip
Inside the background process we make use of \textit{MediaMetadataRetriever} class, to set the location where the video is saved as source path for frames retrieving. Video frames are extracted according to a frame rate, that in here we set to one unit per second. Our decision is justified by trials, considering that a quite slow sliding speed is necessary to take a high-quality video. For the same reason the total number of frames extracted from the video is set to five, as we think they are enough to capture the complete appearance of an object. Anyway these parameters can be tuned as one likes, but it must be carefully considered that a higher number of frames leads to further processing and so to lower speed execution. Each extracted frame is saved locally on the phone through \textit{compress(Bitmap.CompressFormat.JPEG, 100, out)} method, where \textit{out} is a \textit{FileOutputStream} linked to the saving location. When the process of frames extraction is terminated, the \textit{onPostExecute()} method is called, to start predicting their classes.
\bigskip
\begin{lstlisting}
@Override
protected void onPostExecute(Void v) {

  CNNTask cnnTask = new CNNTask(MainActivity.this);
  cnnTask.execute();
}
\end{lstlisting}
\bigskip
The single image classification process, previously shown, in here is extended to all the extracted frames, thus it may take a bunch of seconds to give back the result. Each frame is classified by calling \textit{predictImage()} and asking back the top-5 object classes. For every label we check if it corresponds to the object we are looking for and in that case we save such a frame and  its confidence score. If another frame is found containing our object, after that one is already present, we compare their accuracies and keep the best one, updating the \textit{best\_score} and related Bitmap if the new frame is better.
\begin{lstlisting}
for (int i = 0; i<framePath.length; i++) {
labels = caffeMobile.predictImage(framePath[i],5);
change = false;
  for (int j = 0; j<labels.length; j++){
  if (IMAGENET_CLASSES[labels[j]].contains(insLabel)){
    result = labels[j];
    scores = caffeMobile.getConfidenceScore(framePath[i]);
    if (best_score < scores[labels[j]]){
      best_score = scores[labels[j]];
      change = true;
    }
    j = labels.length;
  }
  }
if (change) {
bmp = BitmapFactory.decodeFile(framePath[i]);
}
}
return result;
\end{lstlisting}
Notice that \textit{framePath} is a String array keeping information about frame paths, as it is needed for classifying them. Finally the frame with the highest probability is returned on the screen, together with the object label, in the very same way as that previously described in \textit{onTaskCompleted} method.
\section{Segmentation module}
In this section we will give a technical insight into the segmentation part. We try to enhance the classification accuracy by further processing the image before predicting its label, trying to separate an object from the background it has around. 
\subsection{Client side}
The user can access this option still from the welcome interface by pressing the proper button. In the fraction of code that handles events coming from buttons pressing, there's a control instruction that checks if the current option is a simple classification or one with segmentation, using an appropriate \textit{segmentation} flag.
\bigskip
\begin{lstlisting}
if (segmentation){
  new HttpAsyncTask().execute();
}
\end{lstlisting}
\bigskip
If the flag value is set to true, we call an asynchronous task that is in charge of performing segmentation. We make use of an asynchronous task because the implementation of segmentation algorithm resides on the web, so the app has to interact with a server to complete its task. Initially the idea was to develop a local solution as it would have been simpler, quicker and safer, but at developing time we faced an issue that couldn't be overcome. In particular we needed OpenCV \cite{culjak2012brief} support to develop a method that relying on image properties would have been able to perform segmentation and we chose to use JavaCPP for that purpose, as it is the best bridge between Java and native C++ code and has also a good support for computer vision libraries. The unsolvable issue that we faced comes from the fact that JavaCPP provides support for 32-bit Android platforms only, while our system is meant to perform on 64-bit platforms, causing in this way an unavoidable clash between the two architectures. That's why we decided to implement the solution in python, which has a nice support for image processing, and put it on a hosting server supporting such libraries. Thus, getting back to our system, it uses an AsyncTask to connect to such a server, sending the image to be processed and getting back its segmentation.
\bigskip
\begin{lstlisting}
class HttpAsyncTask extends AsyncTask<Void, Void, String> {

  @Override
  protected String doInBackground(Void... voids) {
    JSONObject jsonObject;
    bmp = BitmapFactory.decodeFile(imgPath);
    String filename = "filename.png";
    ByteArrayOutputStream bos = new ByteArrayOutputStream();
    bmp.compress(Bitmap.CompressFormat.JPEG, 100, bos);
    byte[] b = bos.toByteArray();
    String encodedImage = Base64.encodeToString
    			  (b, Base64.DEFAULT);
    String result = "";
\end{lstlisting}
\bigskip
Some operations on preparing data need to be done before effectively connecting to the server. A JSONObject is defined as it is needed to encapsulate the image to be sent, then the Bitmap is decoded from the image path and is converted to a byte array. The latter operation is necessary because data have to be transformed in a special String form to be included inside the JSONObject. In particular the byte array is translated into a Base64 string, that is a special way of encoding binary data by translating it into a base 64 representation and returning the new allocated string with the result. Finally \textit{result} variable is declared as it will be the location inside which the segmented image, in string form, will be saved.
\bigskip
\begin{lstlisting}
jsonObject = new JSONObject();
jsonObject.put("imageString", encodedImage);
jsonObject.put("imageName", filename);
String data = jsonObject.toString();
\end{lstlisting}
\bigskip
A new JSONObject is thus initialized and filled with the encoded Base64 version of the image, inserting also its name as it will be useful on server side. Then this object is encoded as a compact JSON string and saved into \textit{data}, making it ready to be sent on the web. After preparing the data, we are ready to start the connection, specifying the proper URL referring to the server location.
\bigskip
\begin{lstlisting}
URL url = new URL(FILE_UPLOAD_URL);
HttpURLConnection urlConnection = url.openConnection();
urlConnection.setRequestMethod("POST");
urlConnection.setRequestProperty("Content-Type",
              "application/json;charset=UTF-8");
OutputStream os = new BufferedOutputStream
        (urlConnection.getOutputStream());
BufferedWriter writer = new BufferedWriter(new
             OutputStreamWriter(os, "UTF-8"));
writer.write(data);
urlConnection.connect();
\end{lstlisting}
\bigskip
\newpage
\textit{FILE\_UPLOAD\_URL} contains the path where the server is hosted, that is \\\textbf{http://loredinho92.pythonanywhere.com/}. We use the \textit{HttpURLConnection} class to define a connection referring to such a URL because it's the most recent way of handling connections in Java, as older ones specifying also the type of request, such as HttpGet and HttpPost, are deprecated. Then we highlight just few of the bunch of connection properties that are set, like \textit{setRequestMethod} that specifies that the request made to the server is a POST, so that data flow both in forward and backward directions, and \textit{setRequestProperty} specifying the character encoding and the content type to be JSON String.
\bigskip
\begin{lstlisting}
if (urlConnection.getResponseCode() == HTTP_OK) {
             
result = readStream(new BufferedInputStream
         (urlConnection.getInputStream()));
         
urlConnection.disconnect();
setResult(result);

return result;
}
\end{lstlisting}
\bigskip
If during the work on the server there's no error, the connection returns an OK code result corresponding to 200 and read the flow of data received from the web through a \textit{BufferedInputStream} defined on top of it, saving the resulting image string inside \textit{result}. In particular its value is retrieved with the following \textit{readStream()} method:
\bigskip
\begin{lstlisting}
private static String readStream(InputStream in) {
  BufferedReader reader = null;
  StringBuilder builder = new StringBuilder();
  try {
        reader = new BufferedReader(new 
        InputStreamReader(in, "UTF-8"));
        String line = "";
        while ((line = reader.readLine()) != null) {
          builder.append(line);
        }
        in.close();
        } catch (IOException e) {
            e.printStackTrace();
        }
  return builder.toString();
}
\end{lstlisting}
\bigskip
The method takes the InputStream defined over the connection as parameter and has as goal that of reading input data flowing from the server and return them as a string. Thus a BufferedReader is defined to read text from a character-input stream, buffering characters so to provide efficient reading of data. In fact each request made by an InputStreamReader causes a corresponding read of the underlying byte stream and if the InputStreamReader wasn't wrapped by a BufferedReader, read operations would cause a considerable level of inefficiency as they are pretty costly. Moreover a \textit{StringBuilder()} is defined to have a structure inside which appending each line read by the BufferedReader. Finally, when the input has no more lines to read, the builder returns its content as a whole string, that in fact is the one referring to the segmented image. Once we have the cropped image, the segmentation phase run in background is basically completed. Thus we call \textit{onPostExecute()} to reconnect with the part of code dealing with simple classification, as we want to predict the label of our new image.
\bigskip
\begin{lstlisting}
@Override
protected void onPostExecute(String result) {
  predict();
}
\end{lstlisting}
\bigskip\bigskip\bigskip
To do this we call \textit{predict()} method that is in charge of reconstructing the Bitmap corresponding the String image, overwriting the segmented image on the location of the old one and starting the AsyncTask that deals with classification.
\bigskip
\begin{lstlisting}
private void predict () {
  byte[] decoded = Base64.decode(result, Base64.DEFAULT);
  bmp = BitmapFactory.decodeByteArray(decoded, 0, decoded.length);
  
  FileOutputStream out=null;
  try {
        out = new FileOutputStream(imgPath);
        bmp.compress(Bitmap.CompressFormat.JPEG, 100, out);
  } catch (Exception e) {
            e.printStackTrace();
  }
    CNNTask cnnTask = new CNNTask(MainActivity.this);
    cnnTask.execute(imgPath);
}
\end{lstlisting}
\bigskip
To recover the Bitmap, image is converted back from Base64 to a byte array, which in turn is given as input to \textit{decodeByteArray()} method of \textit{BitmapFactory} library that gets back the Bitmap object. As mentioned above, reconstructed image needs also to be overwritten on top of the old one, using as usual \textit{FileOutputStream()} class, as its path will be given as input to the classification. Finally the usual background thread meant to predict the image label is started and at the end its result shown on the phone screen as previously described.
\subsection{Server side}
So far we just inspected how the segmentation module works from the client point of view, now we are going to see what happens on the server, showing in detail the image processing steps. First of all the server is hosted by a web hosting service based on Python, whose name is \textbf{pythonanywhere.com} \cite{pythonanywherepythonanywhere}. It is also an online Integrated Development Environment (IDE) providing access to server-based Python and Bash Command-line interface. Moreover it supports Flask \cite{grinberg2014flask}, a Python microframework that we use in order to mange requests coming to the server. The latter is based on the Python WSGI \cite{gardner2009web} utility library Werkzeug and templating language Jinja2 \cite{ronacher2008jinja2}.
\bigskip
\begin{lstlisting}
import base64
from flask import Flask, request
import numpy as np
import cv2

app = Flask(__name__)

@app.route('/', methods=['GET', 'POST'])
\end{lstlisting}
\bigskip
Here is the initial part of the web script, where fundamental libraries are imported, such as \textit{base64} to handle image to String encoding and decoding, \textit{request} to manage the type of the incoming requests, \textit{Flask} as prototype used to create instances of web application and \textit{cv2}, the most important, for OpenCV support in Python \cite{thorne2009introduction}.
Then we need to create an instance of the Flask class for our web app. This is done with \textit{Flask(\_\_name\_\_)}, where \textit{\_\_name\_\_} is a special variable that gets as value the string \textit{"\_\_main\_\_"} when we execute the script. After this initial phase we go to define the main function inside which the incoming requests are managed.
\bigskip
\begin{lstlisting}
@app.route('/', methods=['GET', 'POST'])

def upload_file():
  if request.method == 'POST':
    data = request.json
    img_string = data.get("imageString")
    with open("imageToSave.png", "wb") as fh:
      fh.write(base64.decodestring(img_string))
    segment("imageToSave.png")
    with open("imageToSave.png", "rb") as image_file:
      encoded_string = base64.b64encode(image_file.read())
     return encoded_string
\end{lstlisting}
\bigskip
First we define a function, \textit{upload\_file()}, that at the end returns the encoded string referring to the segmented image. That function is mapped to the home '/' URL, meaning that the client has to navigate to that path to run the script and get the output, and is able to recognize GET and POST requests. Inside its body it checks if the incoming request is a POST, as we are only interested in such a type. If that is the case, sent data are retrieved through the method \textit{request.json}, as the carried payload is a JSONObject string, and saved inside \textit{data}. Then the real string is recovered with \textit{data.get()}, specifying the appropriate field name, decoded with \textit{base64.decodestring()} to reconstruct the image and written on a file that resides on the web. The file path specified in the writing is passed as input to \textit{segment()}, the function that really deals with segmentation, whose result is then converted back into Base64 through \textit{b64encode()} and finally returned to the client. The \textit{segment()} function is the one that really concerns about image processing details.
Its input is represented by the path where the image to be segmented is saved. Initially the image is read with \textit{imread()}, which by default read the image in colored mode and stores its channels in B, G, R order. The first manipulation action that we perform is a noise removal. In particular we use the \textit{GaussianBlur()} method to remove Gaussian noise from the image. A Gaussian kernel must be defined, specifying its width and height in such a way that their values are odd and positive, while the 0 value is related to standard deviation in x and y directions, that in this case is calculated from the kernel size. Then we perform edge detection by using one of the most common methods for such purpose, that is the Sobel operator and running it on each color channel of the image, to finally combine them getting the maximum intensity between the three.
Such an operator computes the difference between gradients along x or y direction. Therefore we need to calculate the Sobel edge on both of them and then obtain the combined magnitude of pixel intensities. This last operation can be executed by using \textit{np.hypot()}, which compute the euclidean distance of each pair of values without the need of any iterative implementation. The next step concerns about noise reduction by computing the mean of the edged image with \textit{np.mean()} and setting to zero every lower intensity value.  
Once we have the edged image with reduced noise, we want to perform contour detection. Contours are represented by curve connecting together all the points along the boundary, which have same intensity, and are very useful in object shape detection. In OpenCV there's a dedicate method to perform this action, \textit{cv2.findContours()}, which take as input the source binary image, the way in which contours should be retrieved and how they must be approximated. The output consists in a list of contours and a hierarchy. Each contour is basically a Numpy array of points which are part of the object boundary. Instead the hierarchy is an array specifying parent-child connections between contours. The idea is that every contour that is enclosed in another one is considered its child and keeps an index referring to the parent. As last step we remove all the contours that do not cover at least a great percentage of the object, getting back only the biggest one.
\section{Database Creation}
The last section is about a quite different visual task, as in here we don't want to classify or process images in anyway. We are rather interested in implementing a solution to what is called the Database Creation problem that could be integrated in the mobile system. Such a problem concerns the fact that training sets are often not very specialized to some particular situation and even if they are really huge, they are still limited and could not scale well to all the needs. A way to overcome this problem could be represented by a simple, quick, light-weight tool that allows anyone to gather object images together with their respective labels. That's why we propose such kind of solution to be integrated as one of the task of our mobile application. The idea is to exploit the smart phone camera to take a scan, or short video, of an object while turning around it, or by using a turning table, to acquire all its relevant views. This is meant to create customized, task-specific databases of annotated images in an easy way, thanks to the phone's portability.
\newpage
In detail when the user decides to access such a function, the phone camera is opened in video mode, waiting for a video capturing. As soon as the video is taken, the asynchronous task responsible for frames extraction is started.
\bigskip
\begin{lstlisting}
@Override
protected Void doInBackground(Void... voids) {
  for (int i = 0; i < 5; i++){
  Bitmap f = mediaMetadataRetriever.getFrameAtTime(i*1000000);
  String timeStamp = new SimpleDateFormat("yyyyMMdd_HHmmss").
				 	  format(new Date());
  framePath[i] = "/storage/emulated/0/Pictures/Caffe-Android/"
          		   +"IMG_"+label+"_"+timeStamp+".jpg";
  try {
        out = new FileOutputStream("/storage
        /emulated/0/Pictures/Caffe-Android/"+
        "IMG_"+label+"_"+timeStamp + ".jpg");
        f.compress(Bitmap.CompressFormat.JPEG, 100, out);
  } catch (Exception e) {
      e.printStackTrace();
      }
  }
  return null;
}
\end{lstlisting}
\bigskip
Frames are extracted according to the previously described frame rate, equal to an unit per second, and for everyone of them a new \textit{timeStamp} is created with \textit{SimpleDateFormat()} to not cause clashes at saving time due to identical names. Each frame path is used to effectively write the frame on the phone drive using \textit{FileOutputStream()}, while \textit{label} is a variable whose content is set according to the input given by the user and refers to the object class for which we are acquiring data.
\chapter{Evaluation and Results}
The network architecture has been trained on an NVIDIA GeForce GTX 1050 with 2 GB of dedicated memory for 30 epochs, using a training and validation batch size of respectively 64 and 32 images. The extracted model has a final accuracy of 45\% and is evaluated on the classification task, considering the simple case and the one with segmentation. The system is also evaluated on object discovery, experimenting two different settings, and on database creation.
\section{Classification on 10-objects database}
First we evaluate the ability of classifying a single image, inspecting different settings in which the object's photo is captured. We assume that the picture's quality is affected by diverse factors, such as: light condition, type of background, distance and view. Light can be natural, as in an outdoor scene, artificial, as from inside a house, or flash so as to light up a condition of darkness. The background can be plain, where the scene rounding the object is basically uniform and no other object is present, or random, where the background can be constituted by anything, accounting for a more real situation, and there also could be other objects occluding the one of interest. As for the distance, the object could be photographed from quite far, so that it doesn't cover a great percentage of the image, or very close, so to put the focus on it. Finally, the image can be taken considering the main view of the object, or from aside, with a more random view resulting from a less precise action. We are interested in understanding which factors affect more the quality of the classification and assume in this way an appropriate guideline when capturing photos so as to enhance the final accuracy.
The results are obtained testing the network architecture on 10 object classes combining together all the aspects previously described. The three tables refer to different light conditions and their entries report success, as a 1, or fail, as a 0, in classifying the object.
According to the experiment the flash light provides a slightly better accuracy, while about the view it is clear that the main view favors a better understanding of the object's nature. Moreover, as we expected, the model has more trouble predicting the object when the picture is taken from far or when the background is not plain. We also observe that the network pays more attention on the object shape with respect to other visual cues, resulting in a better recognition of those objects having a quite complex boundary.

\bigskip\bigskip\bigskip
\begin{figure}[ht]
 \centering
 \captionsetup{justification=centering}
 \includegraphics[scale=0.8]{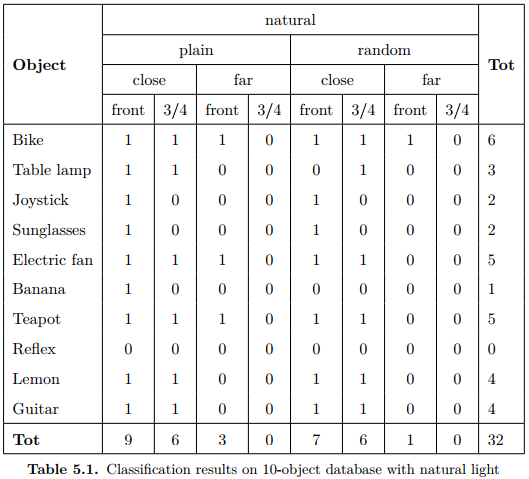}
\end{figure}

\begin{figure}[p]
 \centering
 \captionsetup{justification=centering}
 \includegraphics[scale=0.8]{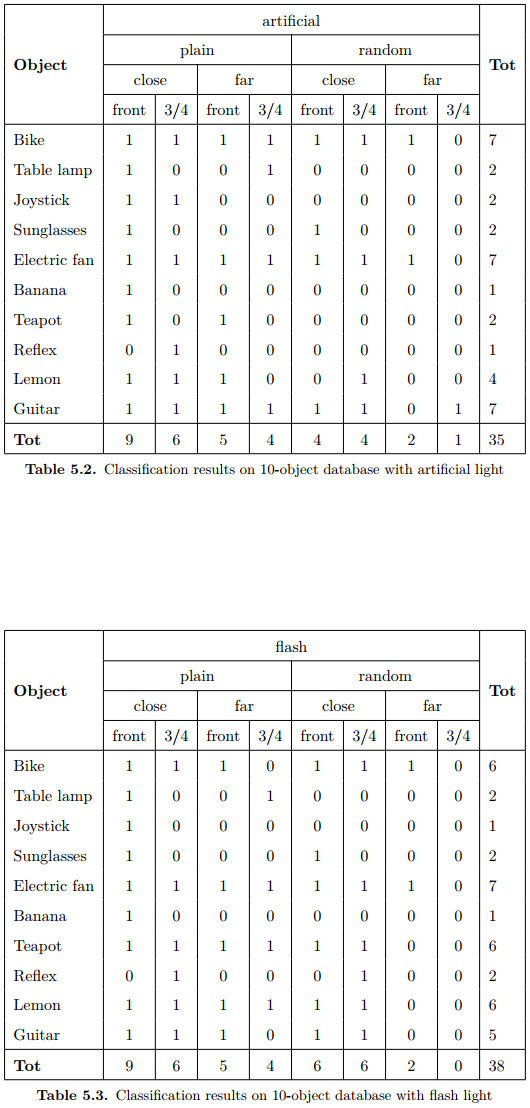}
\end{figure}

\newpage
\section{Classification with segmentation}
Here we show and evaluate the advantage brought by segmenting the object before the effective label prediction. First an important assumption must be made in order to make the algorithm work properly: the image containing the object to be segmented must have a quite plain background and not have other objects whose boundary is in touch with our object's boundary. We now show some example of frames resulting from the segmentation process, considering both an easy case, where the background is relatively uniform, and a harder one where the background is quite cluttered.

\begin{figure}[ht]
 \centering
 \captionsetup{justification=centering}
 \includegraphics[scale=0.45]{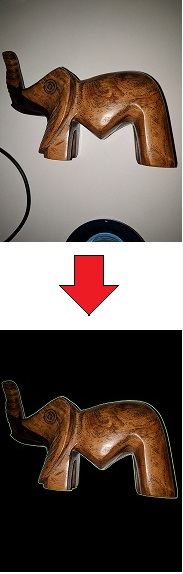}
 \caption{Segmentation example: quite plain background, objects not touching}
\end{figure}

In figure 5.1 segmentation happens very smoothly as in the original image the background is basically plain and the present noise, represented by other object's parts, is not occluding the object of interest.\\
When the background is cluttered as in figure 5.2, the segmentation becomes much harder, even if there's no noise represented by other objects. Anyway the result is still good, as the goal is to reduce the amount of pixels around the object while keeping intact the region occupied by it.

\begin{figure}[ht]
 \centering
 \captionsetup{justification=centering}
 \includegraphics[scale=0.45]{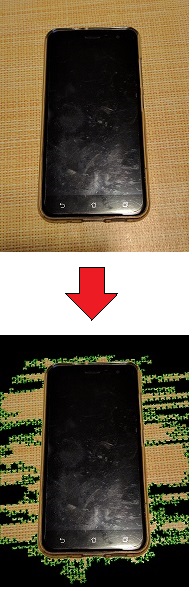}
 \caption{Segmentation example: no other objects, cluttered background}
\end{figure}

\newpage 
Moreover, for each object we take the photo captured in the best combination of factors and test simple classification against classification enhanced with segmentation, showing that for all the considered objects the segmentation step improves the final accuracy.

\bigskip
\begin{figure}[ht]
 \centering
 \captionsetup{justification=centering}
 \includegraphics[scale=0.7]{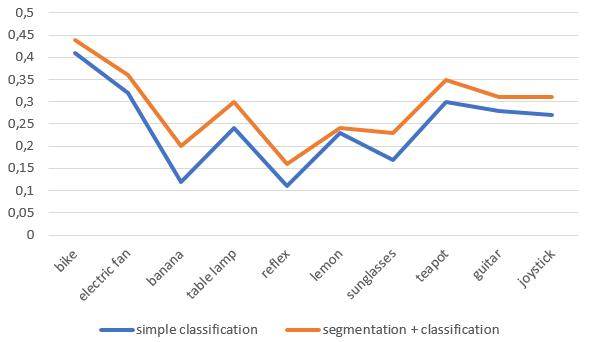}
 \caption{Accuracy on 10 classes: simple classification vs segmentation + classification}
\end{figure}
\newpage
\section{Object Discovery}
Here we test the object discovery task in two typical cases: first we want to simulate the situation in which the robot is looking for a particular object, such as a remote, and acquires short videos of the environment to find it, then we test the system on a scene crowded with several objects, getting back the top-5 detected classes. In the former case, an object label is given as input and the most likely frame is returned as output.

\bigskip \bigskip
\begin{figure}[ht]
 \centering
 \captionsetup{justification=centering}
 \includegraphics[scale=0.85]{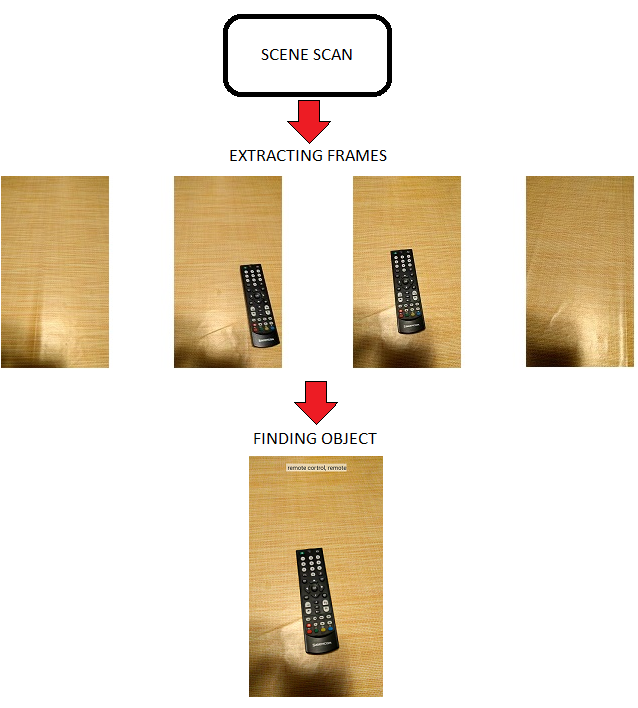}
 \caption{Object discovery: finding a remote in the scene}
\end{figure}

\newpage
In the latter situation no particular label is given as input because we are not looking for a particular object. Instead the idea is to detect more objects possible from the whole scan.

\bigskip \bigskip
\begin{figure}[ht]
 \centering
 \captionsetup{justification=centering}
 \includegraphics[scale=0.4]{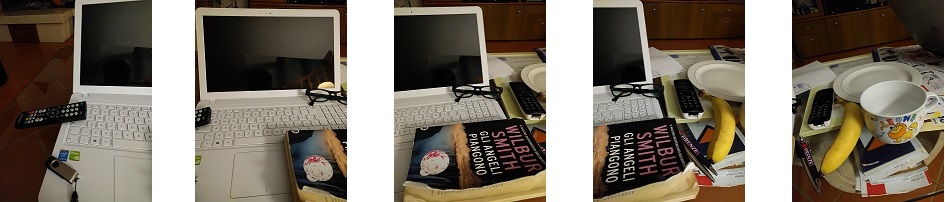}
 \caption{Object Discovery: find top-5 objects in the whole scene}
\end{figure}
\bigskip

In Figure 5.5 the system returns the top-5 objects from the scan of a real scene populated with many different object classes. In particular the detected objects, sorted by accuracy are: \textbf{laptop}, \textbf{screen}, \textbf{keyboard}, \textbf{cup} and \textbf{remote}, confirming that the object that dominates the scene has more probability to be detected.
\section{Database Creation}
In this last section we show an example of data gathering for an object with many different views, such as a laptop.

\bigskip
\begin{figure}[ht]
 \centering
 \captionsetup{justification=centering}
 \includegraphics[scale=0.3]{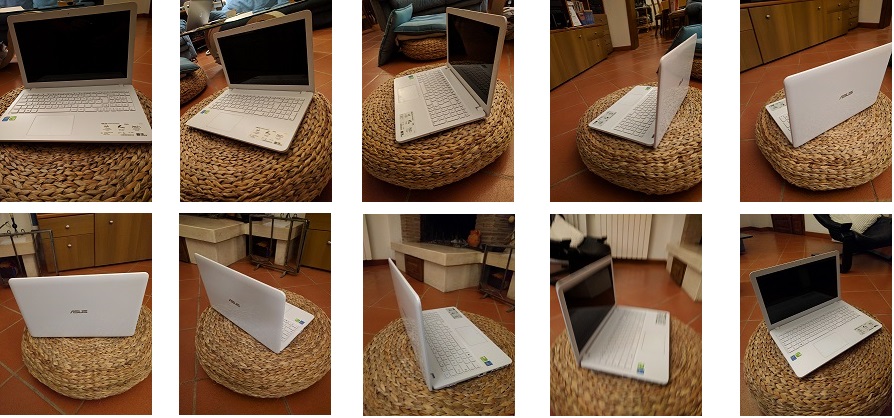}
 \caption{Database Creation: acquiring data from different object views}
\end{figure}
\bigskip

To generate this result both a turntable approach or a simple rotation around the object can be used. All the extracted frames are then saved inside a local database together with the object label.
\chapter{Conclusion}
In this work we have dealt with the development of a mobile application meant to wrap several deep learning tasks in order to provide a scientific tool that allows to test visual models and gather data in a more flexible and portable way, abstracting from the physical robot platform. In particular we have trained the AlexNet deep convolutional architecture on the ILSVRC 2012 and used the extracted model to perform image classification. We have developed a segmentation algorithm, based on OpenCV, to segment the object out of the image before classification and positively demonstrated that this step helps the final label prediction accuracy. We have extended the single classification to a more real situation, where a robot is looking for a certain object and acquires environmental scans to detect it, discretizing the continuous information provided by a video with a frame extraction process and then classifying every single frame to understand where the object is. Finally we have provided a light-weight, high-portable tool to quickly gather annotated visual data taking a short video of an object instance, while slowly rotating around it to capture its main views. Future work will develop in several directions. First, we want to find an alternative way to integrate the segmentation step locally, without the needing of connecting to the server, as well as test different deep learning models. Second, we will investigate some more evolute algorithms to remove image background in such a way that we can lighten the fundamental hypothesis made before segmentation, such as \cite{tommasi2016learning}. Third, we will test different classifiers within this framework, from more powerful architectures always trained over ImageNet to approaches exploiting instead Web-derived data collections \cite{massouh2017learning}, to algorithm explicitly attempting to bridge between the inevitable differences between the images used for training and those found at test time \cite{carlucci2017autodial}.  
\backmatter
\interlinepenalty=1000

\bibliographystyle{unsrt}
\bibliography{biblio}
\begin{acknowledgments}
The project has been developed on a thought of Nizar Massouh. I'd like to thank him for sharing with me the idea of abstracting from the physical constraint represented by the robot platform, to create a more portable and flexible tool, as a mobile app, that could help research in computer vision and visual learning.
\end{acknowledgments}

\end{document}